%% file: main.tex
\documentclass{article}
\usepackage{microtype}
\usepackage{graphicx}
\usepackage{booktabs} 
\usepackage{subcaption}
\usepackage{hyperref}

\newcommand{\Comment}[1]{\quad {\textcolor{gray}{\# #1}}}
\usepackage[accepted]{icml2024}
\input{math_commands.tex}

\usepackage{amsmath}
\usepackage{amssymb}
\usepackage{mathtools}
\usepackage{amsthm}
\usepackage{multirow}
\usepackage{color, colortbl}
\usepackage{wrapfig}
\usepackage{pifont}
\usepackage[capitalize,noabbrev]{cleveref}

\theoremstyle{plain}

\theoremstyle{definition}

\theoremstyle{remark}

\newcommand{\cmark}{\ding{51}}%
\newcommand{\xmark}{\ding{55}}%

\definecolor{Gray}{gray}{0.9}
\definecolor{DarkGray}{gray}{0.75}
\definecolor{darkblue}{rgb}{0,0.08,0.45}
\hypersetup{
  linkcolor = darkblue,
  citecolor  = darkblue,
  colorlinks = true,
}

\usepackage[textsize=tiny]{todonotes}

\icmltitlerunning{Unleashing the Power of Meta-tuning for Few-shot Generalization Through Sparse Interpolated Experts}

\begin{document}

\twocolumn[
\icmltitle{Unleashing the Power of Meta-tuning for Few-shot Generalization \\ Through Sparse Interpolated Experts}



\icmlsetsymbol{equal}{*}

\begin{icmlauthorlist}
\icmlauthor{Shengzhuang Chen}{cityu}
\icmlauthor{Jihoon Tack}{kaist}
\icmlauthor{Yunqiao Yang}{cityu}
\icmlauthor{Yee Whye Teh}{oxford}
\icmlauthor{Jonathan Richard Schwarz}{equal,harvard}
\icmlauthor{Ying Wei}{equal,nanyang}
\end{icmlauthorlist}

\icmlaffiliation{cityu}{City University of Hong Kong}
\icmlaffiliation{kaist}{Korea Advanced Institute of Science and Technology}
\icmlaffiliation{harvard}{Harvard University}
\icmlaffiliation{nanyang}{Nanyang Technological University}
\icmlaffiliation{oxford}{University of Oxford}

\icmlcorrespondingauthor{Shengzhuang Chen}{szchen9-c@my.cityu.edu.hk}
\icmlcorrespondingauthor{Jonathan Richard Schwarz}{schwarzjn@gmail.com}
\icmlcorrespondingauthor{Ying Wei}{ying.wei@ntu.edu.sg}

\icmlkeywords{Machine Learning, ICML}

\vskip 0.3in
]



\printAffiliationsAndNotice{\icmlEqualContribution} 

\begin{abstract}
\input{sections/abstract}
\end{abstract}
\input{sections/intro}
\input{sections/related_works}
\input{sections/preliminary}

\input{sections/method_v2}

\input{sections/experiments}
\input{sections/conclusion}

\bibliography{references}
\bibliographystyle{icml2024}

\newpage
\appendix
\onecolumn

\input{sections/appendix}

\end{document}

%% file: math_commands.tex
\usepackage{amsmath,amsfonts,bm}

\def\eqref#1{equation~\ref{#1}}

\def\1{\bm{1}}

\DeclareMathAlphabet{\mathsfit}{\encodingdefault}{\sfdefault}{m}{sl}
\SetMathAlphabet{\mathsfit}{bold}{\encodingdefault}{\sfdefault}{bx}{n}

\def\gC{{\mathcal{C}}}

\def\gL{{\mathcal{L}}}
\def\gM{{\mathcal{M}}}

\def\gP{{\mathcal{P}}}

\def\gT{{\mathcal{T}}}

\def\gX{{\mathcal{X}}}

\def\sB{{\mathbb{B}}}

\def\sT{{\mathbb{T}}}

\DeclareMathOperator*{\argmin}{arg\,min}

%% file: sections/abstract.tex
Recent successes suggest that parameter-efficient fine-tuning of foundation models is becoming the state-of-the-art method for transfer learning in vision, gradually replacing the rich literature of alternatives such as meta-learning. In trying to harness the best of both worlds, meta-tuning introduces a subsequent optimization stage of foundation models but has so far only shown limited success and crucially tends to underperform on out-of-distribution (OOD) tasks. In this paper, we introduce Sparse MetA-Tuning (SMAT), a method inspired by sparse mixture-of-experts approaches and trained to isolate subsets of pre-trained parameters automatically for meta-tuning on each task. SMAT successfully overcomes OOD sensitivity and delivers on the promise of enhancing the transfer abilities of vision foundation models beyond parameter-efficient fine-tuning. We establish new state-of-the-art results on a challenging combination of Meta-Dataset augmented with additional OOD tasks in both zero-shot and gradient-based adaptation settings. In addition, we provide a thorough analysis of the superiority of learned over hand-designed sparsity patterns for sparse expert methods and the pivotal importance of the sparsity level in balancing between in-distribution and out-of-distribution generalization. Our \href{https://github.com/szc12153/sparse_meta_tuning}{\texttt{code} and \texttt{models}} are publicly available.

%% file: sections/intro.tex
\section{Introduction}\label{sec:intro}




\begin{figure}[t]
\centering
\includegraphics[width=0.975\linewidth]{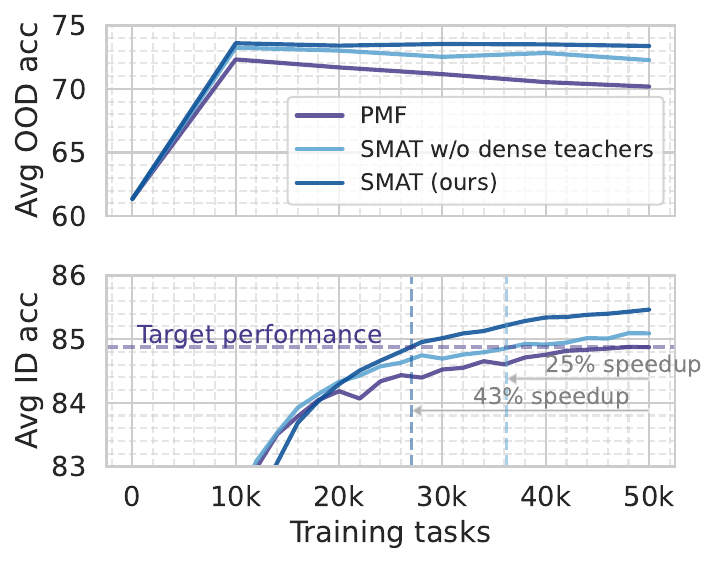}
\vspace{-0.15in}
\caption{Average testing accuracy during meta-training for meta-tuning methods. SMAT yields better ID and OOD results and shows an attractive learning~speedup.}
\label{fig:wallclock_convergence}
\vspace{-0.2in}
\end{figure}
The emergence of foundation models \citep{bommasani2021opportunities} has marked a new chapter in machine learning, with pre-trained models in established domains (e.g., vision or language) becoming virtually indispensable and a vibrant research landscape developing around the design and training of foundation models for new modalities and problems, ranging from the life sciences \citep{lin2023evolutionary} to spectral data \citep{hong2023spectralgpt}, time series \citep{yeh2023toward}, graphs \citep{liu2023towards} and combinations thereof in multi-modal systems \citep{yu2023scaling}. Foundation models have seemingly also signaled the convergence of decades of research on transfer learning (see \citet{zhuang2020comprehensive} for a survey) to the simple yet powerful paradigm of full or parameter-efficient fine-tuning \citep{perez2018film, hu2021lora} of the best foundation model available. As with many breakthroughs in science, this convergence runs contrary to an attractive hypothesis: That the explicit formulation of objectives, algorithms, and optimization procedures targeted directly at downstream performance will result in the best transfer learner (most directly advocated for in the learning to learn, or meta-learning community \citep{thrun2012learning, finn2017model}). Instead, the field's general belief has shifted towards self-supervised objectives such as autoregressive losses \citep[e.g.][]{mikolov2013distributed} or contrastive learning \citep[e.g.][]{radford2021learning} with large models and big data as the best strategy for generalist models with the potential to transfer to a wide variety of tasks.\par

In this paper, rather than committing fully to this view, we instead join a nascent group of researchers aiming to find a middle point between both paradigms, harnessing their strengths and aiming to find a synergy. Indeed, the recent popularity of instruction tuning for large language models \citep{zhang2023instruction} takes a similar view and has emerged as a promising avenue to not only narrow the gap between pre-training and downstream objectives but also enhance zero-shot generalization of pre-trained models. Similarly, \textit{meta-tuning} aims to enhance the transferability of foundation models through a secondary meta-learning stage initiated once pre-training has converged. Indeed, existing research in the field of Natural Language Processing~(NLP) has substantiated the advantages of meta-tuning over traditional fine-tuning and transfer learning approaches, particularly in zero-shot and few-shot testing scenarios~\cite{Gao2021MakingPL,Min2021MetaICLLT,Chen2021MetalearningVL}.\par

Despite initial progress made, the exploration of meta-tuning in vision still remains notably limited to date.~\citeauthor{hu2022pmf} propose a pre-training $\rightarrow$ meta-training $\rightarrow$ fine-tuning pipeline, dubbed PMF~\yrcite{hu2022pmf}, for enhancing the few-shot learning performance of the resulting model relative to the default pre-training $\rightarrow$ fine-tuning approach. With the design principle of simplicity in mind, PMF meta-trains all parameters in a vision transformer using Prototypical Networks~\cite{snell_prototypical_2017} starting from a pre-trained initialization,  yielding the state-of-the-art performance on popular meta-learning benchmarks such as Meta-Dataset~\cite{Triantafillou2020Meta-Dataset}. Despite such promising reported results, we find that this intuitive approach tends to underperform on downstream few-shot tasks, particularly when testing for out-of-distribution~(OOD) tasks (i.e., tasks dissimilar to the ones presented during the meta-training stage).\par

We hypothesize that this low generalization performance on OOD tasks stems from two major factors. (i) The strong emphasis on learning from small amounts of data using a limited number of optimization steps in meta-learning can lead to algorithms that are ``greedy'' w.r.t. the distribution of tasks presented, sacrificing more generalizable features for performance on the distribution at hand. This leads to a risk of meta-overfitting, a phenomenon previously observed \citep[e.g.][]{zintgraf2019fast, yao2021meta, chen2022understanding}. (ii) When meta-tuning tasks are diverse, the default setting of updating all parameters suffers from task interference, making optimization unstable and thereby reducing generalization performance.

In introducing our method, we thus explicitly design core model components to overcome these challenges. We address (i) by taking inspiration from recent work~\cite{Ilharco2022PatchingOM, Panigrahi2023TaskSpecificSL, Wortsman2021RobustFO}, noticing that interpolation between pre-trained and fine-tuned weights leads to a trade-off between ID and OOD generalization performance, with an optimal point usually existing between the extremes. We implement this trade-off through a learned gated interpolation implemented with a sparsity constraint. This particular choice also has the added benefit of addressing (ii) by considering a Mixture-of-Experts inspired approach (with each expert defined through sparse masks), which guarantees expressiveness while alleviating task interference. Finally, sparsity has an additional regularizing effect, further reducing the chance of meta-overfitting and thus counteracting poor OOD generalization observed for standard Meta-Tuning (see Figure \ref{fig:wallclock_convergence} for a direct comparison with the aforementioned PMF). \par

In summary, we propose a reformulation of meta-tuning as a process wherein a hypernetwork undergoes meta-training to select a combination of sparse experts based on few-shot examples, which are subsequently interpolated with the pre-trained model to tailor a powerful foundation model for downstream performance on each specific task.
The integration of an interpolation strategy alongside specialized experts not only preserves the pre-trained model's generalization capabilities but also consolidates the knowledge acquired from all meta-tuned tasks without interference. 
This synergy contributes significantly to our \emph{strong performance across both in-distribution and out-of-distribution few-shot generalization scenarios}. 
Furthermore, we showcase the \emph{interpretability} on task relationship through the experts selected, and \emph{compatibility} of our approach with both full fine-tuning and parameter-efficient fine-tuning methods, such as LoRA~\cite{hu2021lora}.

%% file: sections/related_works.tex
\section{Related Work}

\textbf{Few-shot learning and meta-tuning.}$\ \ $Much of few-shot learning~(FSL) relies on extracting transferable prior knowledge from a collection of few-shot training task episodes through meta-learning~\cite{Hospedales2020MetaLearningIN}, which can then be utilized for data-efficient learning on unseen but related downstream FSL tasks at test time. Meta-learned inductive biases may take the form of a model initialization~\cite{finn2017model}, a learned metric~\cite{snell_prototypical_2017}, a Bayesian prior~\cite{grant2018recasting} or an optimization strategy~\cite{Li2017MetaSGDLT}. A particular subdomain of FSL, namely, cross-domain FSL algorithms~\cite{li2022TaskSpecificAdapter, Triantafillou2020Meta-Dataset, triM, Bateni2019ImprovedFV}, explicitly deals with task-distribution shift between meta-training and -testing. Nevertheless, most architectures used in existing work are limited in scale and without large-scale pre-training. Transitioning into the LLM era, ~\cite{Min2021MetaICLLT} first study meta-training a pre-trained LLM on a large collection of few-shot in-context learning tasks. Their results highlight the effectiveness of meta-training on improving few-shot in-context learning generalization of powerful pre-trained transformers; motivating several follow-up works in the field of meta-tuning in NLP~\cite{Gao2021MakingPL,Min2021MetaICLLT,Chen2021MetalearningVL}. In computer vision, ~\citeauthor{hu2022pmf} propose the simple pre-training, meta-training then fine-tuning (PMF) pipeline~\yrcite{hu2022pmf} and achieve SOTA performance. Concurrent to our work,~\citeauthor{eustratiadis2024neural} explore meta-tuning from an orthogonal direction to ours by proposing a neural architecture search algorithm~\yrcite{eustratiadis2024neural} designed to find the optimal model configuration (e.g., optimal arrangement of adapters) for fine-tuning a meta-tuned model for downstream adaptation.
\par

\textbf{Sparse mixture-of-experts (MoE)}$\ \ $The key idea of sparse MoE is the selective activation of expert modules, usually MLP layers, for each input token during training and inference, thereby achieving graceful scaling. Earlier MoE methods make discrete expert-to-token assignments through a token-choice router scoring experts and selecting the top-k for each token~\cite{shazeer2017,Riquelme2021ScalingVW,lepikhin2021gshard, Fedus2021SwitchTS}. Alternatively, methods may choose the top-k scored tokens for each expert~\cite{zhou2022mixtureofexperts}, use stochastic or fixed routers~\cite{Roller2021HashLF,zuo2022taming}, and more advanced routing techniques~\cite{Lewis2021BASELS,liu2023sparsityconstrained}. However, the discrete nature of the assignment poses a serious challenge to stability in training and optimization~\cite{mustafa2022multimodal, dai-etal-2022-stablemoe}. To this end, more recent works on soft MoE consider a soft relaxation or approximation to the otherwise discrete expert-token assignment~\cite{Puigcerver2023FromST}, as well as other MoE works that employ a weighted sum of experts in the parameter space~\cite{Muqeeth2023SoftMO}.\par

\textbf{Partitioned meta-learning.}$\ \ $The importance of isolating a subset of parameters with high plasticity for optimization-based meta-learning is well-established. A common feature is thus the partitioning of parameters into a set of shared parameters optimized in the outer loop and a (typically smaller) parameter set that implements task adaptation, thus reducing meta-overfitting and memory usage. Early work in this direction \citep[e.g.][]{raghu2019rapid, oh2020boil} rely on partitioning heuristics (such as only updating the last layer) or introduce additional context parameters \citep{zintgraf2019fast} which are concatenated with the input vector. 

Since then, an increasing amount of attention has been placed on adapters, i.e. compact, parameter-efficient modules which have been shown to be particularly impactful when fine-tuning foundation models, thus being particularly suitable for Meta-Tuning. Popular adapter types such as FiLM \citep{perez2018film}, LoRA \citep{hu2021lora} as well as various alternatives feature in various episodic-training methods  \citep[e.g.][]{requeima2019fast, triantafillou2021learning, shysheya2022fit, schwarz2023modality}.

Most closely related to our method is a line of work utilizing sparsification to isolate and train a subset of parameters for rapid adaptation, thus increasing their plasticity for fine-tuning. Most closely related to our method is the aforementioned MSCN \citep{schwarz2022meta}, although the authors focus their experiments on a more specialized compression problem and do not address the problem of how to tackle a specific sparsity level. Alternative approaches feature sparsification in the outer loop through magnitude-based pruning \citep{lee2021meta}, which, while simple, may overly constrain the representational capacity. Similar to over hyper-network inspired approach \citep{schwarz2023modality} predict sparse masks that index into model weights, although they still rely on second-order meta-learning. Finally, the work in \citep{von2021learning} presents a first-order method for gradient sparsity, demonstrating the approach in traditional meta-learning as well as Continual Learning.

%% file: sections/preliminary.tex
\section{Preliminary: Meta-tuning}

Meta-tuning aims to improve the few-show learning performance of a pre-trained model on downstream few-shot testing tasks - usually by directly meta-training the pre-trained model over a collection of labeled few-shot training task episodes ~\cite{hu2022pmf, Min2021MetaICLLT}. Specifically,  we assume the availability of a pre-trained model $f_{\boldsymbol{\theta}^{\mathtt{pre}}}$ as initialization for meta-training, and a training task distribution $\gP_{ID}(\gT)$ from which we may sample fully labeled few-shot training tasks $\gT_i \sim \gP_{ID}(\gT)$. Note that we explicitly denote this as in-distribution ($ID$). In particular, in the~supervised setting, each training task $\gT_i$ takes the form of $\gT_i:=\{\gL_i, \gT_i^s,\gT_i^q\}$, where $ \gL_i$ is the task loss to be minimized, $\gT_i^s :=\{\mathbf{x}_{i,j}^s,\mathbf{y}_{i,j}^s\}_{j=1}^{N_i^s}$ and $\gT_i^q\!:=\!\{\mathbf{x}_{i,j}^q,\mathbf{y}_{i,j}^q\}_{j=1}^{N_i^q}$ are labeled support and query sets of $N_i^s$, $N_i^q$ input-target pairs, respectively. We use the shorthand notations $\mathbf{X}$ and $\mathbf{Y}$ to represent a set of inputs and labels, respectively. Meta-tuning is then realized through the typical episodic-learning setting familiar from meta-learning, i.e., the direct optimization of $\boldsymbol{\theta}$ on the few-shot learning objective which considers minimizing the task loss on the query predictions given information of the support set, i.e., $\boldsymbol{\theta}^{*} :=\argmin_{\boldsymbol{\theta}\in\Theta}\mathbb{E}_{\gP_{ID}}[\gL_i(f_{\boldsymbol{\theta}}(\mathbf{X}^q_i,\gT^s_i),\mathbf{Y}^q_i)]$, over training task episodes sampled from $\gP_{ID}(\gT)$. At test time, we expect to encounter both ID and OOD testing tasks i.e., $\tilde{\gT}:=\{\tilde{\gT}^s_i,\tilde{\mathbf{X}}_i^q\} \sim \gP_{ID}\cup\gP_{OOD}$, where $\gP_{OOD}\neq\gP_{ID}$ is an unseen OOD task distribution. For each testing task, we evaluate the few-shot generalization performance of the meta-tuned model by predicting query labels. Hence, our objective is to develop a meta-tuning algorithm that enables the meta-tuned $\boldsymbol{\theta}^{*}$ to attain optimal few-shot generalization performance across both ID and OOD testing tasks.

%% file: sections/method_v2.tex
\begin{figure*}[t]
  \begin{center}
    \includegraphics[width=0.9\textwidth]{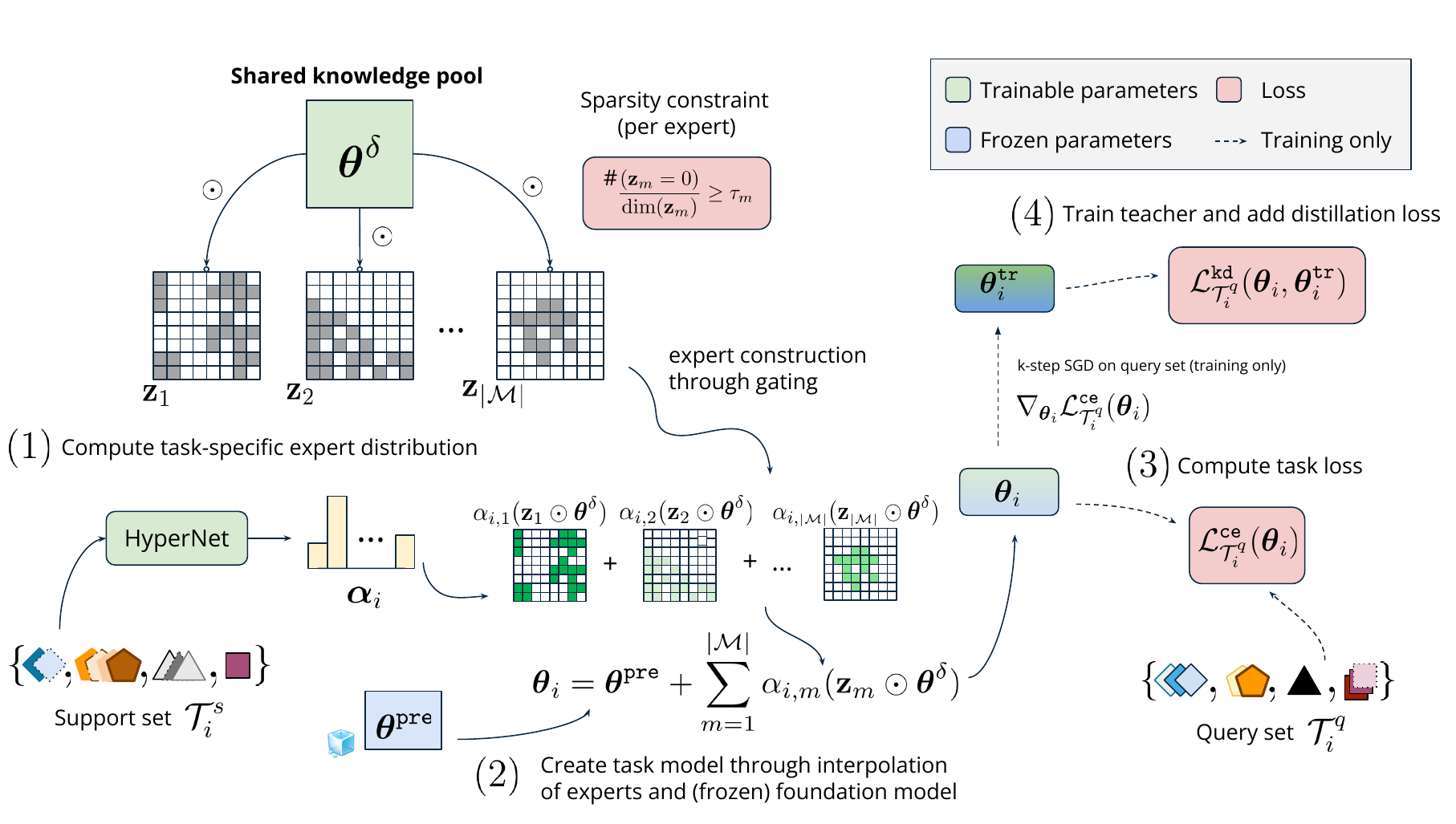}
  \end{center}
  \caption{Overview of the proposed \textbf{S}parse \textbf{M}et\textbf{A}-\textbf{T}uning approach, showing meta-training and inference procedures for a single task $\gT_i$. SMAT meta-learns a shared knowledge pool $\mathcal{M}$ consisting of $|\mathcal{M}|$ sparse interpolated experts characterized by a \textbf{common}, learnable set of dense parameters $\boldsymbol{\theta}^\delta$ and \textbf{distinct}, learnable sets of gating masks $\{\boldsymbol{z}_m\}^{|\mathcal{M}|}_{m=1}$ with sparsity constraints. To construct each task-specific model $\boldsymbol{\theta}_i$ for both meta-training and inference, \textbf{(1)} SMAT first combines experts via a weighted-sum with merging weights $\boldsymbol{\alpha}_i$ generated by a meta-learned hypernetwork $h_{\boldsymbol{\zeta}}$ based on the task's support set $\mathcal{T}_i^s$. \textbf{(2)} The experts are then subsequently combined with the frozen pre-trained model $\boldsymbol{\theta}^\mathtt{pre}$ to enhance both in-distribution (ID) and out-of-distribution (OOD) generalization performance. Alongside \textbf{(3)} the query prediction loss $\gL^\mathtt{ce}_{\mathcal{T}_i^q}(\boldsymbol{\theta}_i)$, \textbf{(4)} knowledge distillation with task-specific dense teachers $\gL^\mathtt{kd}_{\mathcal{T}_i^q}(\boldsymbol{\theta}_i,\boldsymbol{\theta}_i^\mathtt{tr})$ is introduced during meta-training to promote specialization and cooperation of the sparse interpolated experts, ensuring optimization success.}
  \label{fig:overview}
\end{figure*}

\section{SMAT: Sparse MetA-Tuning}
\subsection{Meta-training}
As discussed in~\cref{sec:intro}, naively sharing and updating all pre-trained parameters across all tasks in meta-tuning leads to task interference in optimization~\cite{Yu2020GradientSF, Wang2020GradientVI}. To address this issue, we instead hypothesize that the solution for each task (ID or OOD) comprises a task-specific mixture of a common pool of knowledge covering a broad range of tasks. The knowledge pool is represented by distinct sets of model parameters (i.e., experts), which can be combined cooperatively as a complementary addition to the pre-trained model to promote systematic generalization. Formally, we assume that each task-specific model $\boldsymbol{\theta}_i$ is derived from aggregating the experts via a task-specific weighted sum in the parameter space:
\begin{equation}\label{eq:merge}
    \boldsymbol{\theta}_i = \boldsymbol{\theta}^{\mathtt{pre}} + \sum_{m=1}^{|\gM|}\alpha_{i,m}\boldsymbol{\theta}^{\delta}_m,
\end{equation}
where ${\boldsymbol\theta}_m^{\delta}$ represents the $m$-th expert in the pool $\gM$, and $\boldsymbol{\alpha}_i:=[\alpha_{i,1},\alpha_{i,2},...,\alpha_{i,|\gM|}]$ are the weights.
This way of merging experts in Eqn.~(\ref{eq:merge}) is appealing due to its high expressiveness, as supported by recent findings revealing that merging multiple sets of specialized parameters through simple arithmetic can lead to a better overall multi-task solution~\cite{ilharco2023editing,Matena2021MergingMW}. Moreover, the complementary addition of experts to the pre-trained backbone effectively decouples the two for optimization - meta-training $\{\boldsymbol{\theta}^\delta_m\}^{|\gM|}$ while keeping $\boldsymbol{\theta}^\mathtt{pre}$ frozen provides sufficient capacity and preserves pre-trained knowledge.\par

That said, meta-tuning, in our case, is seen as a procedure for discovering a generalizable selection rule for assigning appropriate experts to tasks. We present an overview of our proposed method in Fig.~\ref{fig:overview}.  

\textbf{Sparsification of experts.}$\ \ $At the core of our method lies the question on \textbf{where} and \textbf{how} to find a set of experts in the model during meta-tuning. Previous studies on partitioned meta-learning, as well as recent works on Mixture-of-Experts, propose a seemingly reasonable solution: Experts are incorporated into a heuristically (and a-priori) chosen \emph{subset} of modules in the model. For example, the batch-norm (BN) layers in ResNets~\cite{Triantafillou2021LearningAU} or the MLP layers in Vision Transformers (ViT)~\cite{Puigcerver2023FromST}. Although these approaches based on fixed partitioning are more memory-efficient and generally result in improved performance compared to their non-expert counterparts, they suffer from a significant bias due to their manually crafted selection (i.e., the concrete choice of $\boldsymbol{\theta}_m^{\delta}$). This bias may be suboptimal when it conflicts with the intricate dynamics of meta-training and, in addition, is not always model-agnostic, e.g., there are no BN layers in ViT.\par

Thus, instead of specifying experts as prior knowledge, we propose to automatically identify the sparsity patterns in experts through meta-tuning using a maximum likelihood objective while considering sparsity constraints on the experts' capacity, thus encouraging specialization. Meta-learning sparsity patterns not only shows minimal bias but also allows for differentiation in the architecture of experts, giving our model a wider range of capabilities to better handle various types of distribution shifts.~\cite{Lee2022SurgicalFI}.\par 

Specifically, we induce a learnable sparsity pattern in the $m$-th expert in the form of a sparse reparameterization $\boldsymbol{\theta}^{\delta}_m\odot\boldsymbol{z}_m$ of the dense expert, where $\boldsymbol{z}_m$ denotes a learnable binary mask with dimension $|\boldsymbol{z}_m|=|\boldsymbol{\theta}^{\delta}_m|$, and $\odot$ denotes element-wise multiplication. 
To learn binary masks through gradient-based optimization, we take inspiration from \citep{schwarz2022meta}, and reparameterize the $m$-th binary mask using an underlying continuous distribution~$q_{\boldsymbol{\phi}_m}$ with parameters $\boldsymbol{\phi}_m$. The reparameterization samples, i.e., $\mathbf{s}_m\!\sim\!q_{\boldsymbol{\phi}_m}(\mathbf{s})$, can be transformed to have values \emph{exactly} $0$ and $1$ through a hard rectification $\mathbf{z}_m\!=\! g(\mathbf{s}_m)\!:=\!\min(1, \max(0, \mathbf{s}_m))$. As a result, sparsity in $\boldsymbol{z}_m$ can be enforced by limiting the probability of $\boldsymbol{s}_m$ being non-zero which can be straight-forwardly expressed through the CDF $Q$ of $q_{\boldsymbol{\phi}_m}$, i.e. $1-Q_{\boldsymbol{\phi}_m}(\mathbf{s}\leq0)$. Choosing $q_{\boldsymbol{\phi}_m}$ as the stretched hard concrete distribution~\citep{louizos2018learning} enables both gradient-based optimization through reparameterization as well as analytical evaluation of the CDF.\par


\textbf{From sparse experts to sparse interpolated experts.}$\ \ $Though the aforementioned sparse reparameterization eventually reduces parameter counts (e.g., by removing zeros in expert parameters), it leads to a strong increase in the number of parameters compared to its non-sparse counterpart at the beginning of meta-tuning -- due to learning $\boldsymbol{z}_m$ and $\boldsymbol{\theta}_m^\delta$ concurrently. Moreover, assigning each~expert its own dense underlying parameters $\boldsymbol{\theta}_m^\delta$ greatly hinders~knowledge transfer among experts, which contradicts the principles of partitioned meta-learning emphasizing collaboration of both task-agnostic and task-specific components.

To tackle these issues, we propose sharing dense parameters across sparse reparameterization of different experts, i.e., $\boldsymbol{\theta}_m^\delta = \boldsymbol{\theta}^\delta, \forall m$. By rearranging Eqn.~(\ref{eq:merge}) slightly, the experts now essentially become different sparse interpolations between the same pre-trained and meta-tuned models~(more details in \cref{app:sparse_interpolations}). Although this formulation of MoE may initially appear bold, it is well-supported by recent works~\cite{panigrahi2023task} 
which suggest that multiple task-specific optimal points can coexist between the same set of pre-trained and meta-tuned models, offering favorable trade-offs for both in-distribution (ID) and some out-of-distribution (OOD) performance. Therefore, our approach of learning sparse interpolated experts can be seen as an inductive bias that promotes the recovery of these optimal interpolation points through meta-tuning. Favorably, this partitioning allows for knowledge transfer among experts through the overlapping regions in their masks. \par

\textbf{Meta-learning expert selection through a hypernetwork.}$\ \ $ In theory, there are two possible approaches to achieve task-specific inference for the expert merging scores $\boldsymbol{\alpha}_i$. The first one is a meta-learned merging score initialization $\boldsymbol{\alpha}_0$ combined with inner-loop gradient-descent on the task support set; the second involves a meta-learned global hypernetwork, $h_{\boldsymbol{\zeta}}(\gT_i^s)$, parameterized by $\boldsymbol{\zeta}$, that directly outputs $\boldsymbol{\alpha}_i$ conditioned on the task support set $\gT_i^s$. We opt for the latter approach as it scales better with larger model sizes. For implementation, we use the pre-trained model $f_{\boldsymbol{\theta}^{\mathtt{pre}}}$ to encode each support image $\mathbf{x}_{i,j}^s\in\mathcal{T}_i^s$ into a vector embedding $f_{\boldsymbol{\theta}^{\mathtt{pre}}}(\mathbf{x}_{i,j}^s)$. The support set embeddings are aggregated into class prototypes which are then concatenated into a sequence and fed into a single trainable transformer block to obtain $\boldsymbol{\alpha}_i'\in\mathbb{R}^{ |\gM| }$ as the output. We treat $\boldsymbol{\alpha}_i'$ as the logits for activating the experts. Instead of choosing the top-k activated ones, which can cause training instability issues and, more importantly, restrict the number of experts per task, we employ the Gumbel-Sigmoid trick~\cite{jang2017categorical} to sample a soft activation value $\in(0,1)$ for each expert, followed by normalization to obtain $\boldsymbol{\alpha}_i\in(0,1)^{|\gM|}$.

\textbf{Enhancing expert specialization through task-specific dense teachers.}$\ \ $Specialization and cooperation among experts play a crucial role in a MoE model. One way to promote specialization is by penalizing the similarity between experts. For example, an orthogonal penalty can be applied to pairs of experts. However, incorporating such explicit penalties makes the optimization problem in the meta-objective more challenging, as it introduces a trade-off with respect to the few-shot prediction performance. To circumvent this trade-off, we propose an alternative approach that utilizes a knowledge distillation loss $\mathcal{L}^\mathtt{kd}_{\mathcal{T}_i^q}(\boldsymbol{\theta}_i,\boldsymbol{\theta}_i^\mathtt{tr})$ \citep{hinton2015distilling} between the merged MoE model $\boldsymbol{\theta}_i$ and a teacher network $\boldsymbol{\theta}_i^\mathtt{tr}$. By using a \emph{highly task-specific} teacher, the distillation loss imposes the weighted-sum of experts that constitute each $\boldsymbol{\theta}_i$ to mimic the behaviour of the teacher, thus enhancing both specialization and cooperation implicitly through knowledge transfer. Furthermore, the predictive performance of $\boldsymbol{\theta}_i$ is not impeded as the loss consistently encourages the current student to achieve a better task-specific generalization on the query set, similar to that of the teacher. \par
We generate the teacher $\boldsymbol{\theta}_i^\mathtt{tr}$ dynamically in each meta-training episode by performing $K$-step gradient descent starting from $\boldsymbol{\theta}_i$ on the query loss $\mathcal{L}^\mathtt{ce}_{\mathcal{T}_i^q}$ (we find $K=1$ is sufficient in practice). It is worth noting that unlike $\boldsymbol{\theta}_i$, we do not limit the capacity of $\boldsymbol{\theta}_i^\mathtt{tr}$ through sparse regularization. This results in a teacher model that is both dense in modulation (i.e., $\boldsymbol{\theta}_{i}^{\mathtt{tr}}-\boldsymbol{\theta}^\mathtt{pre}$), making it expressive and highly task-specific. Importantly, we do not propagate the loss gradients through $\boldsymbol{\theta}_{i}^{\mathtt{tr}}\rightarrow\boldsymbol{\theta}_i$. This approach is thus similar to the use of bootstrapping in optimization-based meta-learning \citep{flennerhag2022bootstrapped, tack2024learning}.\par

\textbf{Meta-optimization with controlled expert sparsity}.$\ \ $We are now ready to state our optimization problem for meta-tuning.
To enable precise control of expert sparsity- which is pivotal for controlling the trade-offs between ID and OOD generalization performance, we solve an optimization problem for few-shot learning performance under sparsity constraints, namely, $ 1 - \frac{L_0(\boldsymbol{z}_m)}{\text{dim}(\boldsymbol{z}_m)}\geq\tau_m, \forall m\in\big[ |\gM| \big]$ where $\tau_m\in[0,1]$ are the targeted sparsity levels. In practice, we optimize the Lagrangian associated with the constraint optimization problem (with Lagrangian multipliers $\boldsymbol{\lambda}$) during meta-tuning:
\begin{align}\label{eq:smat_unconstraint_op}
    \min\limits_{\boldsymbol{\theta}^\delta, \boldsymbol{\zeta}, \boldsymbol{\Phi}} \max\limits_{\boldsymbol{\lambda}\geq \mathbf{0}}\ &\mathbb{E}_{\gT_i\sim \gP_{ID}}\big[\gL^\mathtt{ce}_{\mathcal{T}_i^q}(\boldsymbol{\theta}_i) + \mathcal{L}^\mathtt{kd}_{\mathcal{T}_i^q}(\boldsymbol{\theta}_i,\boldsymbol{\theta}_i^\mathtt{tr})\big] \nonumber \\
    &+ \sum_{m=1}^{|\gM|}\lambda_m(\frac{1}{|\boldsymbol{\phi}_m|}\sum_{k=1}^{|\boldsymbol{\phi}_m|}\tau - Q_{\boldsymbol{\phi}_m}(s_k \leq 0)), \nonumber\\
    \text{where}\quad&\boldsymbol{\theta}_i = \boldsymbol{\theta}^\mathtt{pre} + \sum_{m=1}^{|\gM|}\alpha_{i,m}(\boldsymbol{z}_m\odot\boldsymbol{\theta}^{\delta}), \nonumber\\ 
    &\boldsymbol{z}_m\sim q_{\boldsymbol{\phi}_m}; \boldsymbol{\alpha}_i\sim h_{\boldsymbol{\zeta}}(\gT_i^s),
\end{align}
in which the objective is the minimization problem in the first line while the aforementioned sparsity constraints translate to the maximization problem in the second. We put the sparsity constraints on individual masks, hence experts, as opposed to the overall $\boldsymbol{\theta}_i$ to allow task-dependent model capacity in $\boldsymbol{\theta}_i$ through selective merging. For simplicity, we set a common constraint for all masks i.e., $\tau_m=\tau, \forall m$, which is treated as a hyperparameter. We use simultaneous gradient descent and projected gradient ascent for optimizing Eqn.~(\ref{eq:smat_unconstraint_op}). To avoid over-penalizing the model capacity from surpassing the sparsity constraints, we reset $
\lambda_m$ to zero after its associated sparsity constraint is met~\cite{gallego-posada2022controlled}. This results in final sparsity close to the target $\tau$, enabling precise control of the expert sparsity levels. The pseudocodes for meta-training can be found in~\cref{app:pcode_meta_train}.

\subsection{Meta-testing}\label{sec: mt_adapt}
In the next two sections, we outline the procedures for fine-tuning meta-tuned models using SMAT. SMAT is fully compatible with existing off-the-shelf fine-tuning techniques. Additionally, we introduce a gradient-free fine-tuning method specifically designed for SMAT, which can be optionally employed during downstream task adaptation at test time in more computation-restricted scenarios. Our pseudocodes and ablation results for various fine-tuning techniques with SMAT are provided in~\cref{app:pcode_meta_test} and~\cref{app:ablation_finetune_smat}   .\par
\textbf{Gradient-free optimization for expert selection.}$\ \ $Prior work has discovered that further task-specific adaptation of a meta-trained model is essential for improving its performance on OOD tasks during meta-testing~\cite{hu2022pmf, chen2023secure, li2022TaskSpecificAdapter}. Although effective, one major limitation of these adaptation methods is the reliance on back-propagation of the gradients, which can be expensive, making these methods inefficient and potentially impractical due to poor scaling with model size. To this end, we propose an adaptation strategy, specifically designed for SMAT, that \emph{bypasses the gradient computation}. At the core of our method lies the hypothesis that each expert selection score can be descreatzed i.e, either 0 or 1, which aligns with the intuition that each expert knowledge is either required or not for solving each tasks. We can thus optimize the expert selection score in a binary hypothesize space i.e, $\boldsymbol{\alpha}_i\in\{0,1\}^{|\gM|}$.\par

\textbf{Gradient-based fine-tuning with SMAT.}$\ \ $Following~\cite{hu2022pmf}, we also consider gradient-based fine-tuning of our meta-tuned model at meta-testing time. With only few changes, our method is fully compatible with existing full fine-tuning (i.e., fine-tuning the entire model), and parameter-efficient fine-tuning~(PEFT) techniques. Specifically, we use $\boldsymbol{\theta}_i$ in Eqn.~(\ref{eq:smat_unconstraint_op}) as the task-specific model initialization, before applying any off-the-shelf fine-tuning technique for further optimizing $\boldsymbol{\theta}_i$ on the support set of each task.



%% file: sections/experiments.tex
\section{Experiment}
We now verify the efficacy and competitiveness of SMAT on standard meta-learning benchmarks. 
Additional details and results can be found in~\cref{app:exp_details} and~\ref{app:more_results}.\par

\textbf{Setup.}$\ $We conduct meta-tuning experiments on Meta-dataset~(\textbf{MD})~\cite{Triantafillou2020Meta-Dataset}, which is a widely studied large-scale cross-domain few-shot learning benchmark. As in PMF~\cite{hu2022pmf}, we adhere to the official guidelines and employ the standard meta-training and meta-testing splits for meta-tuning and meta-testing. We select all hyperparameters and the meta-tuned checkpoint for testing using the official meta-validation split. In order to obtain a more comprehensive evaluation of the meta-tuned models, we introduce additional OOD datasets for \emph{meta-testing only}, which were not used during the meta-tuning process on MD.\par

\input{tables/full_results_dino}

\textbf{Baselines.}$\ $We compare SMAT to two types of baselines: \textbf{(a)} the \textbf{Pre}-trained model without meta-tuning,
and \textbf{(b)} meta-tuning methods: \textbf{PMF}~\citep{hu2022pmf}, which is the SOTA on MD. To compare against a MoE baseline for meta-tuning, we adopt the recently proposed SMEAR~\cite{Muqeeth2023SoftMO} which implements soft merging of experts in the parameter space, and denote this baseline by \textbf{SoftMerge}. All methods use DINO-ViT-Small~\cite{caron2021emerging} as the pre-trained backbone.\par


\textbf{Evaluation.}$\ $At meta-testing time, we resort to the ProtoNet~\cite{snell_prototypical_2017} classifier for performing direct inference on each few-shot testing tasks without further adaptation. When considering task-specific fine-tuning on the support sets, we follow the same protocols in PMF~\citep{hu2022pmf} for all models, using \textbf{Full} (fine-tuning the entire model) and \textbf{LoRA}~\cite{hu2021lora} as fine-tuning methods. Namely, for each dataset, we perform a hyperparameter search on a few validation tasks to obtain the optimal learning rate for task-specific fine-tuning using the Adam optimizer for 50 steps for each testing tasks from that dataset.\par


\subsection{SMAT achieves new SOTA performance}
In Tab.~\ref{tab:results_dino}, SMAT consistently achieves the highest overall few-shot classification accuracy for both ID and OOD meta-testing with and without adaptation on task support sets. Results in Table~A\ref{tab:results_sup21k} for supervised pre-trained backbone further demonstrate the superiority of our approach over baselines. More specifically, results in Tab.~\ref{tab:results_dino} show that:\par
\textbf{SMAT is a better out-of-the-box few-shot learner.}$\ \ $SMAT attains the best few-shot learning performance in 5/8 ID and 7/10 OOD datasets without adaptation, outperforming the baseline, PMF, by 0.91\% and 3.17\% on average in ID and OOD evaluation settings, respectively. \par

\textbf{SMAT is a transferable initialization for few-shot fine-tuning.}$\ \ $When considering task-specific adaptation through fine-tuning on the support set of each task, SMAT shows great compatibility with off-the-shelf fine-tuning techniques. Specifically, fine-tuning starting from SMAT's $\boldsymbol{\theta}_i$ leads to the best performance among all baselines when applying the same fine-tuning technique. SMAT improves on PMF by as much as 0.82\% (ID) and 2.48\% (OOD) when fully fine-tuning the entire meta-tuned model as initialization. \par

\textbf{SMAT achieves superior OOD generalization performance.}$\ \ $While PMF exhibits relatively lower OOD performance w.r.t. the pre-trained baselines, SMAT, in contrast, achieves improved generalization performance, outperforming the pre-trained by 3.2\% and at least 0.5\% (when both +LoRA) for without and with adaptation, respectively. \par


\subsection{Roles of sparsity in meta-tuning with SMAT}
\begin{figure}[t!]
    \centering
     \includegraphics[width=0.9\linewidth]{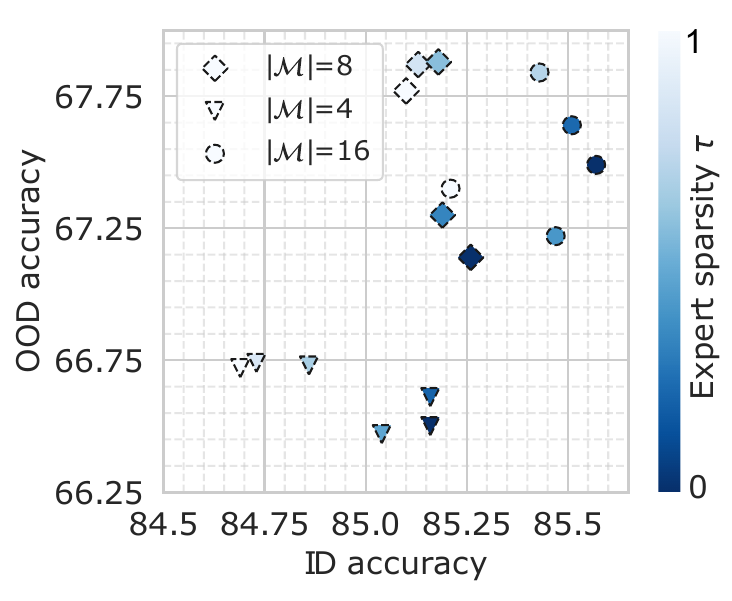}
    \caption{Average performance tradeoff on sampled ID vs OOD tasks as a function of (color) expert sparsity level $\tau$, and (marker) number of experts.}
    \label{fig:sparsity_tradeoff}
\end{figure}

\textbf{ID vs. OOD tradeoff through controlled sparsity levels.}$\ \ $We observe that adjusting the expert sparsity $\tau$ allows us to balance the trade-off between in-domain (ID) and out-of-domain (OOD) performance of our meta-tuned SMAT models. In Fig.~\ref{fig:sparsity_tradeoff}, we see that the OOD performance generally improves while the ID performance decreases as the expert sparsity level $\tau$ increases. We hypothesize that this result is due to the stronger intrinsic meta-regularization effect associated with higher sparsity constraints, as well as the better preservation of the more generic pre-trained features through weight interpolation between the meta-tuned and pre-trained parameters in our formulation in sparse interpolated experts~(see Eqn.~(\ref{eq:smat_unconstraint_op})). Both of these factors help to mitigate meta-overfitting to the ID meta-training tasks hence improve meta-generalization. \par

\begin{figure}[t!]
    \centering
    \includegraphics[width=0.9\linewidth]{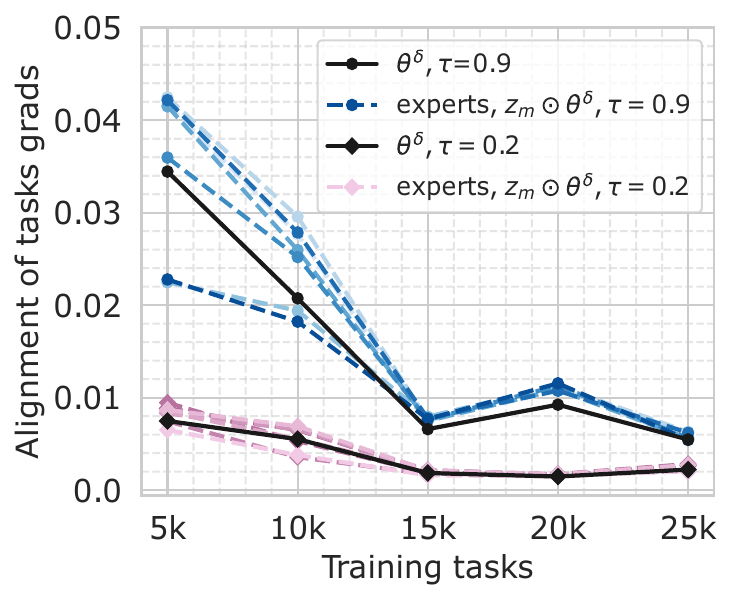}
     \caption{Meta-gradients alignment between tasks throughout for SMAT with low and high sparsity levels. Meta-gradients are calculated w.r.t. the parameters shown in the legend.}
    \label{fig:grad_align_during_training}
\end{figure}

\textbf{Sparsity in experts encourages specialization.}$\ \ $Meta-learned sparsity patterns in the experts~$\boldsymbol{z}_m\odot\boldsymbol{\theta}^\delta$ induce sparsity in the meta-gradients which alleviates harmful task-interference, thereby encouraging specialization among the experts. In Fig.~\ref{fig:grad_align_during_training}, we compare the average alignment of meta-gradients between tasks during training for two SMAT models with different expert sparsity levels $\tau$, where the alignment of gradients, defined as $\mathbb{E}_{\gT_i,\gT_j\sim\gP_{ID}}[\text{cos}(\nabla_{\boldsymbol{\theta}} \gL_i,\nabla_{\boldsymbol{\theta}} \gL_j)]$, is respectively computed for $\boldsymbol{\theta}\in\{\boldsymbol{\theta}^\delta,\boldsymbol{z}_m\odot\boldsymbol{\theta}^\delta,\ \forall m\in[|\gM|]$, i.e., the overall meta-tuned parameter and each expert individually. The results show that the higher sparsity level $\tau=0.9$ in SMAT can lead to greater alignment of meta-gradients between tasks. Moreover, the alignments in the experts' meta-gradients~(which are sparse) are generally higher than that of the overall one, (i.e., w.r.t. $\boldsymbol{\theta}^\delta$ in black) - a sign for development of each expert into highly specialized parameters.
\input{tables/full_results_md_vits}

\begin{figure*}[t!]
\centering
\includegraphics[width=1.0\linewidth]{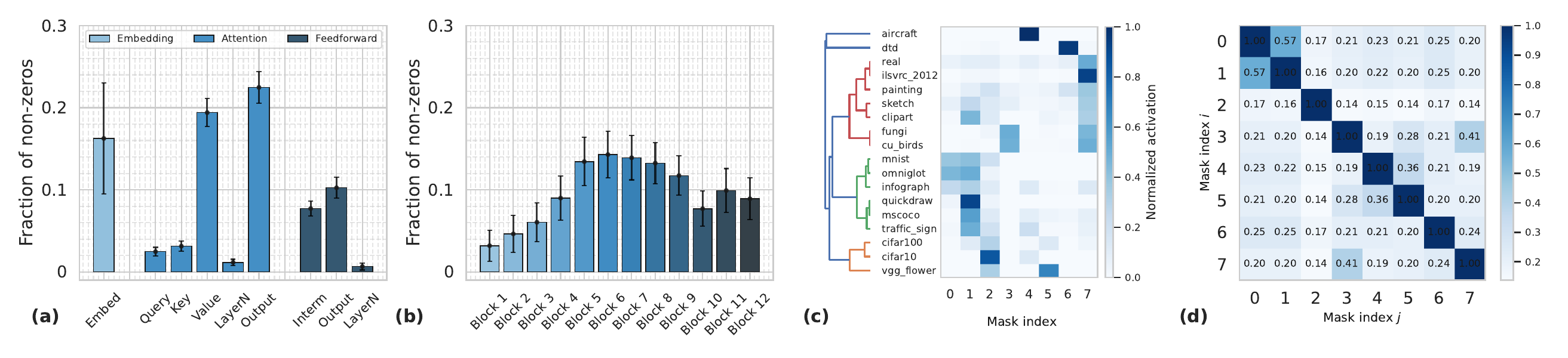}
\vspace{-0.2in}
\caption{(a-b) Meta-learned model capacity after meta-tuning (i.e., number of non-zero parameters) grouped by \textbf{\textit{a)}}: layer types, and \textbf{\textit{b)}}: layer depth. (c-d) Expert specialisation. \textbf{\textit{c)}} Dendrogram of task similarity based on expert selection scores. \textbf{\textit{d)}} Overlap between masks.}
\label{fig:qualitative_visualization}
\vspace{-0.15in}
\end{figure*}

\subsection{Ablation studies}
\textbf{Importance of the different components.}$\ $In Tab.~\ref{tab:ablation_components}, we present several ablated variants of SMAT where we replace or remove certain components.
We perform the study by meta-tuning on the DINO ViT-Small backbone and report the meta-testing results without adaptation. 
Overall, we observe that SMAT performs better than all ablated models on average, demonstrating the effectiveness of each proposed component. We notice that incorporating MoE at the MLP layers of the ViT~(index 5), hence predetermining the sparsity patterns, leads to a marginally better ID performance; however, at the cost of a 1\% drop in its OOD performance compared to SMAT. The results indicate the advantages of explicitly meta-learning the sparsity patterns in the experts for generalization.
\par
\input{tables/ablated_models}

\textbf{Number of experts in SMAT.}$\ \ $In Fig.~\ref{fig:sparsity_tradeoff}, we see that having more experts, hence higher model capacity given the same expert sparsity $\tau$, generally increases both ID and OOD performance of our model. The aforementioned ID vs. OOD tradeoff still exists for different numbers of experts; however, the OOD-to-ID tradeoff ratio~(defined as $\frac{\Delta_{\text{OOD acc}}}{\Delta_{\text{ID acc}}}$) varies - with $|\gM|=4$ experts having the worst tradeoff ratio, and increasing $|\gM|$ from $4\rightarrow8$ leads to the most significant gain in the ratio while the improvement seems to saturate when further increasing $|\gM|$ from $8\rightarrow16$.\par

\begin{figure}[h!]
    \centering
    \includegraphics[width=0.9\linewidth]{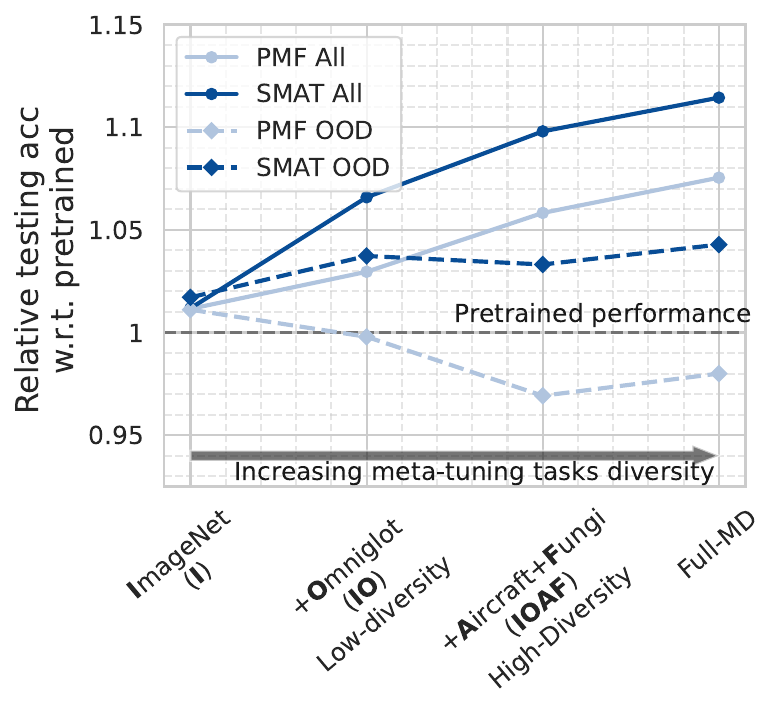}
    \caption{Relative testing performance w.r.t. the pre-trained initialization ($\frac{\text{Avg acc meta-tuned}}{\text{Avg acc pre-trained}}$) for SMAT vs PMF using meta-training tasks of increasing diversity.}
    \label{fig:md_subsets}
\end{figure}

\textbf{Scale of Meta-tuning datasets.}$\ \ $In Fig.~\ref{fig:md_subsets}, we investigate the impact of the quantity and diversity of tasks observed during meta-tuning on the generalization performance of meta-tuned models on unseen meta-testing tasks. As anticipated, the overall generalization performance (on ID and OOD testing tasks), of both PMF and our model, improves as the scale of the meta-tuning datasets increases along the x-axis. However, even with increased quantity and diversity in the meta-tuning tasks, the OOD performance of PMF is not always better than that of the pre-trained model, which we conjecture is due to both meta-overfitting and harmful task interference. 
In contrast, SMAT consistently achieves 
better OOD and overall performance compared to both PMF and pre-trained models, with a noticeable $\sim$4\% improvement in relative OOD performance even in the low-diversity (IO) scenario. It is worth noting that this is also where the largest improvement in OOD performance for SMAT occurs. Intuitively, this is because Omniglot is very different from ImageNet, which was the only training source prior to its addition, resulting in a significant increase in task diversity. With only these two datasets, SMAT achieves comparable OOD performance to its full-MD version. These results highlight the effectiveness of using SMAT for meta-tuning in low data diversity settings, as well as its ability to achieve improved few-shot generalization by better leveraging task diversity during meta-tuning.

\textbf{Scale of Vision Transformer backbones.}$\ $In Tab.~\ref{tab:results_different_vits}, we present the meta-testing results on MD of various Vision Transformer backbones meta-tuned with SMAT. Our findings indicate that the larger models tend to offer superior overall performance but may require substantially more computational resources due to the number of parameters. These results can guide the selection of appropriate transformer architectures based on specific application requirements and resource constraints.

\subsection{Qualitative visualization}
\textbf{Patterned sparsity emerges through meta-tuning.}$\ \ $In Fig.~\ref{fig:qualitative_visualization}, we visualize the sparsity patterns on masks identified through meta-tuning. We observe that the sparsity levels vary significantly depending on the layer types (a), and depths (b). Specifically, the intermediate layers (5-9) have lower per-layer sparsity, while the first few layers are highly sparsified, with sparsity levels as high as 95\%. Among the different layer types, we find that three types of layers retain most of their modulation parameters (non-zeros): (1) the first input embedding layers, (2) values of the attention module, and (3) linear layers of attention and feedforward modules. Across different masks, we notice that the standard deviations of sparsity levels are particularly larger for the first embedding layers and throughout layers at all depths. By examining the overlapping ratio~(defined as $\frac{|(\boldsymbol{z}_i\cap\boldsymbol{z}_j)\neq0|}{|(\boldsymbol{z}_i\cup\boldsymbol{z}_j)\neq0|}$), as shown in Fig.~\ref{fig:qualitative_visualization}(d), we find that different masks, hence experts, generally have a small overlap. This indicates that SMAT has indeed discovered a diverse set of sparse interpolated experts through meta-tuning.\par

\textbf{Learned expert merging rule encodes task relationship.}$\ \ $A closer look at the average expert selection scores $\boldsymbol{\alpha}$ by datasets reveals that both specialized experts and a meaningful selection rule have been meta-learned by SMAT, as evident in the Fig.~\ref{fig:qualitative_visualization}(c). We first note that overall, every expert has been utilized by some domains. More interestingly, the dendrogram produced by the similarity of mask selection scores clearly shows hierarchical clustering according to visual similarities between domains. Furthermore, we note sparsity and discreteness of expert selection generally inversely correlates with tasks complexity, with more sparse and discrete selection for intuitively simpler tasks~(e.g., Omniglot, Dtd) than the more complex ones~(e.g., Ilsvrc\_2012).

%% file: tables/full_results_dino.tex
\begin{table*}[t!]
  \centering
  \caption{Few-shot testing results on the Meta-dataset benchmark and additional OOD testing datasets for methods using DINO-ViT-Small backbone. $^\dagger$ and $^\ddagger$ respectively indicate published results in $^\dagger$\cite{hu2022pmf} and $^\ddagger$\cite{basustrong2023}. \colorbox{Gray}{Gray} indicates our method.}
\resizebox{\textwidth}{!}{
\begin{tabular}{lccccccccccc}
    \multicolumn{1}{c}{} & \multicolumn{4}{c}{w/o fine-tuning} & \multicolumn{7}{c}{with gradient-based fine-tuning} \\
    \cmidrule(r){1-1}\cmidrule(lr){2-5}\cmidrule(lr){6-9}\cmidrule(l){10-12}
     Datasets    & $^\dagger$Pre &$^\dagger$PMF & SoftMerge& \cellcolor{Gray}SMAT  & $^\ddagger$Pre+full & $^\dagger$PMF+full & SoftMerge+full & \cellcolor{Gray}SMAT+full & $^\ddagger$Pre+LoRA & PMF+LoRA & \cellcolor{Gray}SMAT+LoRA  \\
   \cmidrule(r){1-1}\cmidrule(lr){2-5}\cmidrule(lr){6-9}\cmidrule(l){10-12}
    ImageNet & 73.48 & 73.54 & 74.33 &  \cellcolor{Gray}\textbf{74.69}     & 73.54 & 74.59 & 74.71&\cellcolor{Gray}\textbf{75.24} & 74.22 & 73.54 & \cellcolor{Gray}\textbf{75.72}\\ 
    Aircraft & 62.17 & 88.33 & 88.80 &  \cellcolor{Gray}\textbf{89.78}     & 75.4  & 88.33 & \textbf{90.60}&\cellcolor{Gray}{90.01} & 80.8  &   89.75 &	\cellcolor{Gray}\textbf{90.71}\\
    Omniglot & 54.33 & \textbf{91.79} & 91.24 & \cellcolor{Gray}89.84      & 78.7  & 91.79 & \textbf{92.01}&\cellcolor{Gray}90.83 & 80.8  &    \textbf{92.78}	&\cellcolor{Gray}90.99\\
    CUB   & 85.37 & 91.02 & 91.54 &  \cellcolor{Gray}\textbf{92.57}   & 85.4  & 91.02 & 91.95&\cellcolor{Gray}\textbf{92.57} & 85.8  &   91.17	&\cellcolor{Gray}\textbf{92.57}\\
    DTD   & 83.67 & 81.64 & 80.98& \cellcolor{Gray}\textbf{86.29}       & 86.9  & 86.61 & 86.84&\cellcolor{Gray}\textbf{88.41} & 86.8  &    86.73	&\cellcolor{Gray}\textbf{88.28}	\\
    Quickdraw & 60.59 & \textbf{79.23} & 78.98 &  \cellcolor{Gray}79.17     & 73.6  & 79.23 & \textbf{79.90}&\cellcolor{Gray}79.17 & 72.7  &     \textbf{79.23}	&\cellcolor{Gray}78.83\\
    Fungi & 56.26 & \textbf{74.2}  & 72.40  &  \cellcolor{Gray}73.31     & 54.7  & \textbf{74.20}  & 72.40&\cellcolor{Gray}73.31  & 59.8  &      \textbf{75.44}	&\cellcolor{Gray}73.31\\
    VGGFlower & 94.45 & 94.12 & 96.89 &  \cellcolor{Gray}\textbf{97.22}     & 94.2  & 94.12 & 97.01&\cellcolor{Gray}\textbf{97.22} & 94.8  &      96.05&	\cellcolor{Gray}\textbf{97.25}\\
     \cmidrule(r){1-1}\cmidrule(lr){2-5}\cmidrule(lr){6-9}\cmidrule(l){10-12}
    \textbf{ID Avg} & {71.29} & {84.23} &  {84.40} & \cellcolor{Gray}\textbf{85.36} & {77.81} & {84.99} & 85.56&\cellcolor{Gray}\textbf{85.84} & {79.47} & {85.59} & \cellcolor{Gray}\textbf{85.88} \\
     \cmidrule(r){1-1}\cmidrule(lr){2-5}\cmidrule(lr){6-9}\cmidrule(l){10-12}
    TrafficSig & 53.7  & 54.37 & 56.21&  \cellcolor{Gray}\textbf{57.72}      & 87.3  & 88.85 & 89.91&\cellcolor{Gray}\textbf{91.33} & 88.1  &     89.14	&\cellcolor{Gray}\textbf{90.18}\\
    MSCOCO & 54.58 & 57.04 & 55.75&   \cellcolor{Gray}\textbf{58.81}     & 61.5  & 62.59 & 62.15&\cellcolor{Gray}\textbf{63.11} & 62.1  &      61.71	&\cellcolor{Gray}\textbf{63.38}\\
    Cifar10 & 85.64 & 80.82 & 84.58 &  \cellcolor{Gray}\textbf{87.05}     & \textbf{92.48} & 89.61 & 91.84&\cellcolor{Gray}92.21 & \textbf{93.33} &     91.53	&\cellcolor{Gray}92.46\\
    Cifar100 & 76.86 & 69.11 & 70.85 &  \cellcolor{Gray}\textbf{77.46}     & \textbf{86.13} & 82.54 & 85.88&\cellcolor{Gray}86.12 & \textbf{86.17} &     85.06	&\cellcolor{Gray}85.88\\
    MNIST & 78.57 & 93.33 & 94.16 &  \cellcolor{Gray}\textbf{94.43}     & 92.54 & 96.44 & 96.20&\cellcolor{Gray}\textbf{96.73} & 94.98 &    96.41	&\cellcolor{Gray}\textbf{96.46}\\
    Sketch & 47.25 & 41.10  & 43.30 &  \cellcolor{Gray}\textbf{47.76}     & 56.39 & 49.65 & 53.85&\cellcolor{Gray}\textbf{56.67} &\textbf{57.34} & 47.59 & \cellcolor{Gray}55.63\\
    Pet& 91.73 & 91.37 &89.84& \cellcolor{Gray}\textbf{91.97} &    \textbf{92.03} & 91.73 & 90.48&\cellcolor{Gray}91.97 & 92.06 & 92.01&\cellcolor{Gray}\textbf{92.31} \\
    Clipart & 55.19 & 53.92 &  54.83&  \cellcolor{Gray}\textbf{58.97}   & \textbf{67.18} & 62.83 & 65.50&\cellcolor{Gray}65.79 & \textbf{66.51} & 60.6  &\cellcolor{Gray}66.07 \\
    Food & 62.64 & 61.89 &63.04& \cellcolor{Gray}\textbf{65.59} &     65.08 & 62.97 & 63.36&\cellcolor{Gray}\textbf{66.99} & 65.06 & 62.71 & \cellcolor{Gray}\textbf{67.77} \\
    Cars& 34.58 & \textbf{38.00} & 36.21 & \cellcolor{Gray}36.79&  40.98 & 40.07 & 41.62&\cellcolor{Gray}\textbf{42.39} & 39.49 & \text{42.37} & \cellcolor{Gray}{40.05} \\
      \cmidrule(r){1-1}\cmidrule(lr){2-5}\cmidrule(lr){6-9}\cmidrule(l){10-12}
    \textbf{OOD Avg} & {64.07} & {64.10} & {64.87} & \cellcolor{Gray}\textbf{67.65} & {74.16} & {72.73} & 74.08&\cellcolor{Gray}\textbf{75.32} & {74.51} & {72.91} & \cellcolor{Gray}\textbf{75.02} \\
      \cmidrule(r){1-1}\cmidrule(lr){2-5}\cmidrule(lr){6-9}\cmidrule(l){10-12}
\end{tabular}
}
  \label{tab:results_dino}
\end{table*}

%% file: tables/full_results_md_vits.tex
\begin{table*}[h!]
  \centering
  \caption{Few-shot testing results on the Meta-dataset benchmark for various SMAT meta-tuned Vision Transformer backbones. }
\resizebox{\textwidth}{!}{
\begin{tabular}{cllllllllllllllll}
\multicolumn{3}{c}{}&\multicolumn{14}{c}{with gradient-based full fine-tuning} \\
\toprule
\multicolumn{2}{l}{\multirow{2}{*}{\textbf{SMAT meta-tuned ViT backbone} }} & \multirow{2}{*}{\textbf{Params}.}&\multicolumn{9}{c}{\textbf{In-domain}}                                            & \multicolumn{2}{l}{\textbf{Out-of-domain}} &          &       & \multirow{2}{*}{\textbf{Avg}}      \\
\cmidrule(r){4-11}\cmidrule(lr){12-16}
\multicolumn{2}{l}{} & & INet  & Acraft & Omglot & CUB   & DTD   & Qdraw & Fungi & Flower & Sign  & Coco           & Cifar10          & Cifar100 & Mnist &    \\
\cmidrule(r){1-2}\cmidrule(lr){3-3}\cmidrule(lr){4-11}\cmidrule(lr){12-16}\cmidrule(lr){17-17}
\multicolumn{2}{l}{DINO Small~\cite{caron2021emerging}}         &       21M                     & 75.24 & 90.01  & 90.83  & 92.57 & 88.41 & 79.17 & 73.31 & 97.21  & 91.33 & 63.11          & 92.08            & 85.91    & 96.73 & 86.42 \\
\multicolumn{2}{l}{DINO Base~\cite{caron2021emerging}}            &    86M                          & 78.28 & 92.19  & 92.89  & 93.41 & 86.97 & 80.35 & 75.59 & 96.93  & 91.23 & 64.85          & 92.57            & 86.95    & 97.70  & 87.22 \\
\multicolumn{2}{l}{Sup21k Small~\cite{Dosovitskiy2020AnII}}                             &       21M    & 82.10 & 89.16  & 88.56  & 94.98 & 89.05 & 79.36 & 74.50 & 99.19  & 90.13 & 68.19          & 94.87            & 90.32    & 96.76 & 87.81 \\
\multicolumn{2}{l}{Sup21k Base~\cite{Dosovitskiy2020AnII}}                             &     86M         & 85.01 & 87.41  & 88.05  & 95.52 & 87.59 & 79.78 & 74.01 & 99.34  & 91.38 & 68.51          & 93.61            & 92.84    & 96.88 & 87.96 \\
\multicolumn{2}{l}{Sup21k Large~\cite{Dosovitskiy2020AnII}}                               &  307M       & 87.38 & 88.00     & 88.16  & 95.59 & 88.56 & 81.02 & 74.34 & 99.09  & 91.77 & 69.50          & 95.27            & 92.62    & 96.79 & 88.43\\
\bottomrule
\end{tabular}
}
\label{tab:results_different_vits}
\end{table*}

%% file: tables/ablated_models.tex
\begin{table}[H]
  \caption{Ablation studies on different components of SMAT. \textbf{MLS}: meta-learned sparsity, \textbf{Meta}: Meta-training using support and query splits (otherwise no split), \textbf{DT}: dense teachers. \textbf{IE}: interpolated experts}
  \small
  \centering
  \label{tab:ablation_components}
  \begin{sc}
\resizebox{0.95\linewidth}{!}{
   \begin{tabular}{llccccccc}
    \toprule
    ID &Model& MLS& Meta& DT & IE& ID & OOD & Avg\\
    \cmidrule(r){1-2} \cmidrule(lr){3-6}\cmidrule(l){7-9}
    1&SMAT&\cmark&\cmark&\cmark&\cmark&85.14&\textbf{67.27}&\textbf{75.21}\\
    2&&\cmark&\cmark&\xmark&\cmark&85.07& 66.44&74.74\\
    3&&\cmark&\cmark&\cmark&\xmark&84.77&67.02&74.90\\
    4&&\cmark&\xmark&\cmark&\cmark& 82.35&63.64&71.95\\
    5&&\xmark&\cmark&\cmark&\xmark&\textbf{85.21}&66.21&74.75\\
    \cmidrule(r){1-2} \cmidrule(lr){3-6}\cmidrule(l){7-9}
    6&PMF &\xmark&\xmark&\xmark&\xmark&84.23& 64.09& 73.05\\
    \bottomrule
  \end{tabular}
}
  \end{sc}
\end{table}

%% file: sections/conclusion.tex
\section{Conclusion}
We introduced a simple-yet-effective meta-tuning framework coined SMAT that accommodates to each task through an interpolation of the pre-trained model and a learned combination of sparse experts. Our experiments conclusively demonstrate SMAT's effectiveness in delivering a more generalizable pre-trained model, resulting in state-of-the-art performance on out-of-distribution datasets. Notably, SMAT seamlessly integrates with cutting-edge parameter-efficient fine-tuning methods, and analyses of sparsity patterns underscore the specialization of the learned experts.

\section*{Acknowledgement}
This work was supported by the Research Matching Grant Scheme (RMGS 9229111) founded by the University Grants Committee of Hong Kong, and the Institute of Information \& communications Technology Planning \& Evaluation (IITP) grant funded by the Korea government (MSIT) (No.2022-0-00713, Meta-learning applicable to real-world problems)

\section*{Impact Statement}
The proposed meta-tuning strategy exhibits a broad applicability, poised for extension or adaptation beyond vision pre-trained models. We envision this project as pioneering a new research trajectory for improving few-shot generalization of foundational models across diverse domains, including natural language processing, life sciences, time series, and more.

By augmenting few-shot generalization in pre-trained models, particularly in real-world applications often situated outside of the pre-training distribution, the proposed meta-tuning approach stands poised to substantially impact various downstream tasks like medical imaging analysis, self-driving, wildlife monitoring and etc.

%% file: sections/appendix.tex
\input{sections/appendix/pcode}

\newpage
\newpage
\input{sections/appendix/exp_details}
\input{sections/appendix/additional_results}

\input{sections/appendix/discussion}


%% file: sections/appendix/pcode.tex
\section{Pseudocode for SMAT}

\subsection{SMAT for meta-training}\label{app:pcode_meta_train}
The pseudocode for meta-tuning using SMAT can be found in Alg.~\ref{alg:smat_metatraining} below. Our implementation is publicly available at \href{https://github.com/szc12153/sparse_meta_tuning}{\texttt{github.com/szc12153/sparse\_meta\_tuning}}.
\par


\begin{algorithm}[h!]
\caption{SMAT: Meta-training}
\label{alg:smat_metatraining}
\begin{algorithmic}

 \STATE {\bfseries Data:} \text{Meta-training tasks} $\sT:=\{\gT_1,\gT_2,...\gT_N\}\sim\mathcal{P}_{ID}$\\
\STATE {\bfseries Require:} \text{Pre-trained initialization} $\boldsymbol{\theta}^\mathtt{pre}$; \text{Target expert sparsity level} $\tau_m=\tau$; \text{Number of experts} $M$\\
\STATE {\bfseries Output:} \text{Variational distribution} $\Phi=\{\phi_m\}^{|\gM|}_{m=1}$ \text{for the sparse masks}; \text{Hypernetwork} $h_\psi$; \text{Meta-tuned parameters} $\boldsymbol{\theta}^\delta$

\STATE Initialize $\zeta,\ \Phi,\ \theta^\delta$
\STATE \text{Initialize the Lagrangian multipliers $\lambda_m=0$ for sparsity constraint $1 -\frac{L_0(\boldsymbol{z}_m)}{\text{dim}(\boldsymbol{z}_m)}\geq\tau,\ \ \forall m\in[|\gM|]$.}
\WHILE{\text{not converged}}
    \STATE $\sB\sim\sT$  \Comment{Sample a batch of tasks}
    \FOR{$i =1,2,...,|\sB|$}
        \STATE $\gT_i \rightarrow \gT^s_i, \gT_i^q$ \Comment{Split into support and query sets}
        \STATE $\boldsymbol{\alpha_i'}\sim \mathtt{GumbleSigmoid}(h_\zeta(\gT^s_i))$ \Comment{Sample expert merging scores from the hypernetwork}
        \STATE $\alpha_{i,m} = \frac{\alpha_{i,m}'}{\sum_m^{M}\alpha_{i,m}'}$ \Comment{Normalize the merging weights}
        \STATE $\boldsymbol{z}_m\sim q_{\phi_m},\ \forall m\in[M]$ \Comment{Sample sparse masks from the variational distribution}
        \STATE $\boldsymbol{\theta}_i = \boldsymbol{\theta}^\mathtt{pre} +  \boldsymbol{\theta}^{\delta}\odot\sum_{m}^{M}\alpha_{i,m}\boldsymbol{z}_m$ \Comment{Weighted-sum of sparse experts}
        
        \STATE $\boldsymbol{\theta}^\mathtt{tr}_i\leftarrow\mathtt{StopGrads}(\boldsymbol{\theta}_i)$
        \FOR{$k=1,2,...,K$ }
            \STATE{\Comment{Task-specific dense teacher}}
            \STATE $\boldsymbol{\theta}^\mathtt{tr}_i \leftarrow \mathtt{GradientDescent}(\nabla_{\boldsymbol{\theta}^\mathtt{tr}_i}\gL_{\gT_i^q}^\mathtt{ce}(\boldsymbol{\theta}^\mathtt{tr}_i))$
        \ENDFOR
        \STATE $\gL_i := \beta \gL^\mathtt{ce}_{\mathcal{T}_i^q}(\boldsymbol{\theta}_i) + (1-\beta)\mathcal{L}^\mathtt{kd}_{\mathcal{T}_i^q}(\boldsymbol{\theta}_i,\boldsymbol{\theta}_i^\mathtt{tr})$ \Comment{task's meta-loss, $\beta$ is a weighting coefficient $\in (0,1)$} 
    \ENDFOR
    \STATE $\gC o_m:=\frac{\lambda}{\text{\small dim}(\boldsymbol{\phi}_m)}\sum_{j}^{\text{dim}(\boldsymbol{\phi}_m)}Q_{\boldsymbol{\phi}_m}(s_j\leq0) - \tau,\ \forall m\in[M]$ \Comment{sparsity loss for each expert}

    \STATE $[\boldsymbol{\zeta},\boldsymbol{\Phi}, \boldsymbol{\theta}^\delta]\leftarrow\mathtt{GradientDescent}(\nabla_{[\boldsymbol{\zeta},\boldsymbol{\Phi}, \boldsymbol{\theta}^\delta]}\big(\frac{1}{|\sB|}\sum_{i=1}^{|\sB|}\mathcal{L}_i+\sum_{m=1}^M\lambda_m\gC o_m)\big)$ \Comment{Minimization problem in Eqn.~(\ref{eq:smat_unconstraint_op})}
    \FOR{$m=1,2,...,M$} 
    \IF{$\gC o_m<0$ }
        \STATE \Comment{expected expert sparsity is lower than the constraint $\tau$}
        \STATE $\lambda_m\leftarrow  \mathtt{GradientAscent}(\nabla_{\lambda_m}\gC o_m\frac{\sum_{i\in\mathbb{B}}\alpha_{i,m}}{|\mathbb{B}|})$ \Comment{Maximization problem in Eqn.~(\ref{eq:smat_unconstraint_op})}
    \ELSE{}
    \STATE \Comment{expected expert sparsity is at least $\tau$, the constraint is satisfied
    \STATE $\lambda_m\leftarrow 0$ } \Comment{Reset the Lagrangian multiplier for the $m$-th constraint}
    \ENDIF 
    \ENDFOR

\ENDWHILE
\STATE \textbf{Return} $\boldsymbol{\Phi},\boldsymbol{\zeta}, \boldsymbol{\theta_\delta}$
\end{algorithmic}
\end{algorithm}

\subsection{SMAT for meta-testing}\label{app:pcode_meta_test}
We propose a heuristic in Alg.~\ref{alg:gradient_free_expert_selection} for optimizing the task-specific expert selection during meta-testing time without the need for expensive gradient computation. Specifically, we restrict the normalized expert selection score estimated by our hypernetwork $h_\zeta$ in binary states i.e., $\alpha_{i,m}' \in \{0, 1\},\ \forall m\in [M]$, and optimize $\alpha_{i,m}$ in this binary space by minimizing the loss on the meta-testing task's support set. The intuition behind this is that each expert is either needed or discarded for each few-shot learning tasks, which is supported by our empirical observations on $\alpha_{i,m}'$ being very discrete (close to 0 or 1) in most cases.\par

\begin{algorithm}[h!]
   \caption{SMAT: \textbf{Meta-testing time gradient-free expert selection} (for a single task $\tilde\gT_i$)}
   \label{alg:gradient_free_expert_selection}
\begin{algorithmic}
 \STATE {\bfseries Input:} Testing support set $\tilde\gT_i^s$ and query inputs $\tilde{\mathbf{X}}_i^q$, Meta-trained $\zeta,\theta^\delta,\Phi$, Pre-trained $\theta^\mathtt{pre}$
   \STATE $\boldsymbol{\alpha}_i'\sim \mathtt{HardGumbleSigmoid}(h_\zeta(\tilde\gT^s_i))$ \Comment{Initialize expert merging scores using the hypernetwork and round to [0,1]}
   \STATE $l*$ = positive infinity \Comment{use to record the lowest support loss during exploration}
   \STATE $\boldsymbol{z}_m\sim q_{\phi_m},\ \forall m\in[M]$ \Comment{Sample sparse masks once at the start }
    \FOR{$r = 1,2,...,R$ }
    \STATE \Comment{repeat for R rounds of sampling}
        \FOR{$m = 1,2,...,M$}
        \STATE \Comment{iterate through each score in $\alpha_i$}
        \STATE Flip $\alpha_{i,m}',\ 0\leftrightarrow1$ \Comment{Generate candidate score for the $m$-th expert}
        \STATE $\alpha_{i,m} = \frac{\alpha_{i,m}'}{\sum_m^{M}\alpha_{i,m}'}$ \Comment{Normalize the merging weights}
        \STATE $\boldsymbol{\theta}_i = \boldsymbol{\theta}^\mathtt{pre} +  \boldsymbol{\theta}^{\delta}\odot\sum_{m}^{M}\alpha_{i,m}\boldsymbol{z}_m$ \Comment{Weighted-sum of sparse experts}
        \STATE $\tilde\gL_i := \gL^\mathtt{ce}_{\tilde\gT_i^s}(\boldsymbol{\theta}_i)$ \Comment{Evaluate the support loss which only requires forward passes}
        \IF{$\tilde\gL_i<l$ } 
        \STATE \Comment{Rejection sampling}
        \STATE Accept the candidate $\alpha_{i,m}$ with $\rho$ and record $l*=l=\tilde\gL_i$, otherwise reject
        \ELSE{}
        \STATE Accept the candidate $\alpha_{i,m}$ with $(1-\rho)$ and record $l=\tilde\gL_i$ otherwise reject
        \ENDIF 
         \ENDFOR
     \ENDFOR
    \STATE  {\bfseries Return:} $\boldsymbol{\alpha_i}$ at the lowest support loss $l^*$, Which is then used for final prediction on the query $\hat{\mathbf{Y}}^{q}_i=f(\tilde{\mathbf{X}}_i^q;\boldsymbol{\theta}_i = \boldsymbol{\theta}^\mathtt{pre} +  \boldsymbol{\theta}^{\delta}\odot\sum_{m}^{M}\alpha_{i,m}\boldsymbol{z}_m$.
\end{algorithmic}
\end{algorithm}

\begin{algorithm}[h!]
   \caption{SMAT: \textbf{Meta-testing time full fine-tuning} using $\theta_i$ as an model initialization (for a single task $\tilde\gT_i$)}
   \label{alg:collapse_then_full}
\begin{algorithmic}
\STATE {\bfseries Input:} Testing support set $\tilde\gT_i^s$ and query inputs $\tilde{\mathbf{X}}_i^q$; Meta-trained $\zeta,\theta^\delta,\Phi$, Pre-trained $\theta^\mathtt{pre}$ 
    \STATE $\boldsymbol{\alpha}_i'\sim \mathtt{GumbleSigmoid}(h_\zeta(\tilde\gT^s_i))$ \Comment{Obtain expert merging scores using the hypernetwork}
    \STATE $\alpha_{i,m} = \frac{\alpha_{i,m}'}{\sum_m^{M}\alpha_{i,m}'}$ \Comment{Normalize the merging weights}
        \STATE $\boldsymbol{\theta}_i = \boldsymbol{\theta}^\mathtt{pre} +  \boldsymbol{\theta}^{\delta}\odot\sum_{m}^{M}\alpha_{i,m}\boldsymbol{z}_m$ \Comment{Weighted-sum of sparse experts}
        \STATE $\boldsymbol{\theta}_{i,0} \leftarrow\mathtt{StopGrads}(\boldsymbol{\theta}_i)$
        \FOR{$k=0,2,...(K-1)$} 
            \STATE $\boldsymbol{\theta}_{i,k+1}\leftarrow\mathtt{GradientDescent}(\nabla_{\boldsymbol{\theta}_{i,k}}\gL_{\tilde\gT_i^s}^\mathtt{ce}\boldsymbol{\theta}_{i,k})$ \Comment{finetune $\theta_{i}$ on the support set for $K$ steps}
        \ENDFOR
\STATE  {\bfseries Return:} Final prediction on the query set $\hat{\mathbf{Y}}^{q}_i=f(\tilde{\mathbf{X}}_i^q;\boldsymbol{\theta}_{i,K})$
\end{algorithmic}
\end{algorithm}



%% file: sections/appendix/exp_details.tex
\section{Experiment details}\label{app:exp_details}
\subsection{Implementation of baselines}
\textbf{PMF}~\cite{hu2022pmf}.$\ \ $In Tab.~\ref{tab:results_dino}, we report published results by PMF in their paper~\cite{hu2022pmf} whenever they are available. For extra meta-testing datasets in Tab.~\ref{tab:results_dino}, which were not included in their paper, we produce these results using the official meta-trained PMF model checkpoint, which is publicly available with their code on Github. 


\subsection{Details for meta-testing}
\textbf{W/O fine-tuning}.$\ \ $To perform direct few-shot inference for a given testing task, $\tilde{\gT}_i:=\{\tilde{\mathbf{X}}_i^s, \tilde{\mathbf{Y}}^s_i, \tilde{\mathbf{X}}_i^q\}$, we consider using the prototypical network~\cite{snell_prototypical_2017}, which first constructs class centroids in the feature space of the model using the labeled support set, before performing the nearest centroid classification on the query input. \par
Denote the feature backbone (e.g., a pre-trained or meta-tuned ViT) by $f_{\boldsymbol{\theta}}:\gX\rightarrow\mathbb{R}^{D}$ parameterized by $\boldsymbol{\theta}$, which essentially is a mapping from the input pixel space to a vector space of dimension $D$. The centroid for each unique class $\mathbf{c}_k, k=[1,2,3...K]$ in $\tilde\gT_i$ is calculated from the support set by: $$\mathbf{c}_k = \frac{1}{|j:y_{i,j}^s=c_k|}\sum_{j:y_{i,j}^s=c_k}f_{\boldsymbol{\theta}}(\mathbf{x}_{i,j}^s).$$ To this end, the predicted label for each query input is given by $$p(y=k|\mathbf{x}^q) = \frac{\text{exp}(-d(f_{\boldsymbol{\theta}}(\mathbf{x}^q), \mathbf{c}_k))}{\sum_{k'=[1,2..]}^{K}\text{exp}(-d(f_{\boldsymbol{\theta}}(\mathbf{x}^q), \mathbf{c}_k'))},$$ where $d(\cdot,\cdot)$ is some distance metric e.g., cosine distance.


%% file: sections/appendix/additional_results.tex
\section{Additional experimental results} \label{app:more_results}
\subsection{Meta-dataset meta-testing results for meta-tuned supervised pre-trained backbones}
Table~\ref{tab:results_sup21k} provides the results of the meta-tuned models on MD using PMF~\cite{hu2022pmf} and SMAT with Sup21K-ViT-Small backbone. \par 
\textbf{SMAT attains the best performance.}$\ \ $ Although using a different pre-trained backbone, we again observe that SMAT outperforms all baselines in both ID and OOD meta-testing. These results, together with our main results in Tab.~\ref{tab:results_dino}, validate the efficay of our approach.\par

\textbf{Comparing to few-shot meta-testing results for meta-tuned DINO backbone.} Meta-tuning on self-supervised and supervised pre-trained backbones produces vastly different generalization results on OOD tasks. While for both PMF and ours, meta-tuning with the Sup21K backbone generally improves few-shot testing performance over the pre-trained backbone even on unseen tasks and domains, meta-tuning with a self-supervised pre-trained backbone (e.g., DINO) requires taking more care in the design of meta-tuning strategy - noticeably, naively meta-tuning with PMF can be outperformed by simple pre-trained + fine-tuned baselines, particularly on OOD tasks; in contrast, SMAT, still maintains the high transferability in its meta-tuned feature representation which leads to better, if not better, fine-tuning performance when using ours meta-tuned model as the fine-tune initialization.\par 
\input{tables/full_results_sup21}

\subsection{Different fine-tuning strategies for SMAT at meta-testing}~\label{app:ablation_finetune_smat}

\textbf{Performance.}$\ \ $In Tab~\ref{tab:ablation_finetune} below, we evaluate various meta-testing fine-tuning strategies for SMAT and compare their performance. We first note the effectiveness of our proposed gradient-free expert selection method (see~\cref{sec: mt_adapt}), as evidenced by its improved performance compared to directly using SMAT with ProtoNet~\cite{snell_prototypical_2017} for meta-testing. Second, using $\theta_i$ as the initialization for full fine-tuning, which has the same capacity as the pre-trained model $\theta^\mathtt{pre}$, leads to improved performance over fine-tuning the full SMAT model jointly (i.e., \textbf{(4)}), as it is sufficiently expressive while much more parameter-efficient than the latter hence avoids potential over-fitting issues. We provide more explaination as follows.\par

Gradient-based fine-tuning outperforms gradient-free fine-tuning primarily because it is much more flexible. The extra flexibility of gradient-based fine-tuning (Alg. 3, Appendix) stems from the fact that the entire model $\theta_i$ is allowed to be updated, whereas gradient-free fine-tuning (Alg. 2, Appendix) only allows updates on the expert selection weights $\alpha_i$ - with dimensions equal to the number of experts in SMAT; while all other parameters remain frozen. That said, the hypothesis space of our gradient-free fine-tuning algorithm is much more tightly constrained around the model meta-tuned on ID tasks. As a result, the effectiveness of gradient-free fine-tuning is limited on certain OOD tasks that exhibit noticeable distribution shift, such as TrafficSign. In such cases, the meta-tuned ID model may become inadequate, requiring significant parameter updates.\par
\begin{table}[h!]
  \caption{Different fine-tuning strategies for SMAT on a subset of few-shot testing tasks. \textbf{(1)}: Direct inference by ProtoNet, reported in~Tab.~\ref{tab:results_dino} as SMAT; \textbf{(2)}: Gradient-free expert selection (Alg.~\ref{alg:gradient_free_expert_selection}); \textbf{(3)} Full fine-tuning using $\theta_i$ as initialization (Alg.~\ref{alg:collapse_then_full}), reported in~Tab.~\ref{tab:results_dino} as SMAT+Full. \textbf{(4)}: Full fine-tuning the entire SMAT model, i.e., $\boldsymbol{\zeta}, \boldsymbol{\theta^{\delta}}, \boldsymbol{\Phi}$ jointly.}
  \tiny
  \centering
  \label{tab:ablation_finetune}
  \begin{sc}
\resizebox{0.6\linewidth}{!}{
   \begin{tabular}{lcccc}
    & \multicolumn{2}{c}{{Gradient-free}} & \multicolumn{2}{c}{{Gradient-based}} \\
    \cmidrule(r){1-1}\cmidrule(lr){2-3}\cmidrule(l){4-5}
    Datasets & (1) & (2)  & (3)  & (4)\\
    \cmidrule(r){1-1}\cmidrule(lr){2-3}\cmidrule(l){4-5}
    TrafficSign&58.51&59.59&90.83&89.93\\
    MSCOCO&57.35&58.78&63.07&62.76\\
    Cifar10&83.95&86.95&92.08&91.97\\
    Cifar100&74.85&77.20&85.91&85.95\\
    MNIST&94.53&94.63&96.73&96.70\\
    \cmidrule(r){1-1}\cmidrule(lr){2-3}\cmidrule(l){4-5}
  \end{tabular}
}
  \end{sc}
\end{table}

\textbf{Computational cost.}$\ \ $ We present results below in Tab.~\ref{tab:ablation_finetune_cost} for a quantitative comparison of computational cost, in terms of time and GPU memory, between w/o fine-tuning, gradient-free fine-tuning, and gradient-based full fine-tuning for SMAT. All fine-tuning are carried out in FP16 mixed precision. In particular, our gradient-free fine-tuning method offers a $+1.59$\% improvement on average over w/o fine-tuning at no additional memory cost, while saves $3/4$ of the total memory cost of gradient-based fine-tuning. Gradient-based fine-tuning, however, outperforms both by at least $+10.31$\% despite requiring 4x the GPU memory of both, and 2x the time cost compared to gradient-free fine-tuning. 

\begin{table}[h!]
  \caption{Computational cost for fine-tuning a SMAT model (ViT-DINO-Small backbone) at meta-testing time with different fine-tuning strategies.}
  \tiny
  \centering
  \label{tab:ablation_finetune_cost}

\resizebox{0.8\linewidth}{!}{
   \begin{tabular}{lccc}
    Method & Time (sec./task) & GPU memory (MiB) & Avg. Acc (Tab.~\ref{tab:ablation_finetune})\\
    \cmidrule(r){1-1}\cmidrule(lr){2-4}
    w/o fine-tuning (ProtoNet)& 0.2 & 4332 & 73.83 \\
    Alg.(~\ref{alg:gradient_free_expert_selection}). gradient-free fine-tuning  (50steps)&  6.4  & 4332& 75.43 \\
    Alg.(~\ref{alg:collapse_then_full}). gradient-based full fine-tuning (50steps)&11.8 & 17264 &85.74\\
    \cmidrule(r){1-1}\cmidrule(lr){2-4}
  \end{tabular}}

\end{table}

\subsection{Performance vs number of parameters.}~\label{sec:performance_vs_parameter_counts}
Details on parameter counts: SMAT: We use a very naive compression scheme to remove exact zeros in our model. We use a $(\mathtt{value},\mathtt{position})$ tuple to represent each non-zero parameter in our model after flattening all parameters in a single long vector. Thus, the total number of parameters left in the experts is equal to two-times the number of non-zero parameters remained at the end of meta-tuning. We point out that there are perhaps more memory efficient ways for representing sparse weights e.g., PyTorch sparse tensors,  which could potentially result in a more significant saving in terms of number of binary \emph{bits}.\par
\begin{figure}[h!]
\centering
\includegraphics[width=.45\linewidth]{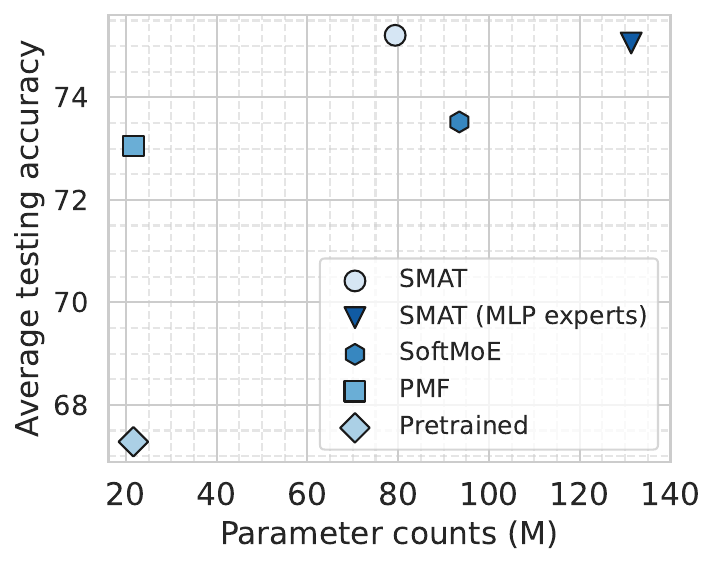}
\vspace{-5mm}
\caption{Average testing performance of models vs model parameter counts. We observe SMAT attains the best overall performance while requiring less number of parameters comparing to other MoE implementation variants including SoftMoE~\cite{Puigcerver2023FromST}, and incorporating experts in the MLP layers in ViT (SMAT MLP experts).}
\label{fig:performance_vs_num_params}
\end{figure}

%% file: tables/full_results_sup21.tex
\begin{table}[htbp]
  \centering
   \caption{Few-shot testing results on the Meta-dataset benchmark and additional OOD testing datasets for methods using Sup21k-ViT-Small backbone~\cite{Dosovitskiy2020AnII}. SMAT indicates our method.}
   \resizebox{0.85\textwidth}{!}{
    \begin{tabular}{lccccccccc}
          & \multicolumn{3}{c}{Gradient-free} & \multicolumn{6}{c}{Gradient-based} \\
    \cmidrule(r){1-1}\cmidrule(lr){2-4}\cmidrule(l){5-10}
    Meta-dataset & \multicolumn{1}{c}{Pre} & \multicolumn{1}{c}{PMF} & \multicolumn{1}{c}{Ours} & \multicolumn{1}{c}{Pre+full} & \multicolumn{1}{c}{PM+full} & \multicolumn{1}{c}{SMAT+full} & \multicolumn{1}{c}{Pre+LoRA} & \multicolumn{1}{c}{PMF+LoRA} & \multicolumn{1}{c}{SMAT+LoRA} \\
    \cmidrule(r){1-1}\cmidrule(lr){2-4}\cmidrule(l){5-10}
    ImageNet & 68.45 & 79.57 & \textbf{81.39} & 78.96 & 80.37 & \textbf{82.10} & 76.30 & 80.21 & 81.89 \\
    Aircraft & 52.57 & 85.55 & \textbf{87.05} & 83.20 & 88.48 & \textbf{89.16} & 80.73 & 87.40 & 88.40 \\
    Omniglot & 37.02 & \textbf{86.63} & 86.14 & 78.25 & \textbf{88.58} & 88.56 & 73.75 & 88.05 & 87.91 \\
    CUB   & 84.65 & 94.72 & \textbf{94.98} & 91.20 & 94.72 & \textbf{94.98} & 90.18 & 94.72 & \textbf{94.98} \\
    DTD   & 80.75 & 84.84 & \textbf{86.27} & 87.20 & 88.60 & \textbf{89.05} & 86.59 & 87.85 & 88.50 \\
    Quickdraw & 55.05 & 78.55 & \textbf{79.35} & 75.75 & \textbf{80.01} & 79.36 & 74.07 & 78.55 & 79.35 \\
    Fungi & 44.20 & 73.02 & \textbf{74.50} & 56.29 & 73.02 & \textbf{74.50} & 55.88 & 73.02 & 74.50 \\
    VGGFlower & 94.11 & 98.97 & \textbf{99.09} & 98.01 & 99.09 & \textbf{99.19} & 96.97 & 99.06 & 99.14 \\
    \cmidrule(r){1-1}\cmidrule(lr){2-4}\cmidrule(l){5-10}
    \textbf{Avg ID}   & 64.60 & 85.23 & \textbf{86.10} & 81.11 & 86.61 & \textbf{87.11} & 79.31 & 86.11 & \textbf{86.83} \\
    \cmidrule(r){1-1}\cmidrule(lr){2-4}\cmidrule(l){5-10}
    TrafficSig & 48.14 & 55.80 & \textbf{60.10} & 90.02 & \textbf{90.13} & \textbf{90.13} & 89.04 & 88.89 & 89.52 \\
    MSCOCO & 52.39 & 63.77 & \textbf{63.91} & 64.64 & 67.02 & \textbf{68.19} & 64.08 & 67.12 & 67.65 \\
    Cifar10 & 79.33 & 87.50 & \textbf{91.10} & 93.40 & 93.78 & \textbf{94.87} & 92.80 & 92.31 & 93.62 \\
    Cifar100 & 68.53 & 79.11 & \textbf{82.02} & 88.54 & 88.81 & \textbf{90.32} & 87.69 & 88.81 & 88.80 \\
    MNIST & 73.53 & 93.90 & \textbf{94.01} & 95.16 & 96.70 & \textbf{96.76} & 94.82 & 96.51 & 96.67 \\
    \cmidrule(r){1-1}\cmidrule(lr){2-4}\cmidrule(l){5-10}
    \textbf{Avg OOD} & 64.38 & 76.02 & \textbf{78.23} & 86.35 & 87.29 & \textbf{88.05} & 85.69 & 86.76 & \textbf{87.25} \\
    \cmidrule(r){1-1}\cmidrule(lr){2-4}\cmidrule(l){5-10}
    \end{tabular}%
    }
  \label{tab:results_sup21k}%
\end{table}%

%% file: sections/appendix/discussion.tex
\section{Sparse interpolated experts}\label{app:sparse_interpolations}
As previously stated, \textit{by rearranging Eqn.~(\ref{eq:merge}) slightly, the experts now essentially become different sparse interpolations between the same pre-trained and meta-tuned models}. Here are the details.\par
Starting from Eqn.~(\ref{eq:merge}), we have:
\begin{align}
     \boldsymbol{\theta}_i &= \boldsymbol{\theta}^{\mathtt{pre}} + \sum_{m=1}^{|\gM|}\alpha_{i,m}\boldsymbol{\theta}^{\delta}_m\\
     & = \boldsymbol{\theta}^{\mathtt{pre}} + \sum_{m=1}^{|\gM|}\alpha_{i,m}(\boldsymbol{z}_m\odot\boldsymbol{\theta}^{\delta})\\ & = \boldsymbol{\theta}^{\mathtt{pre}} + \sum_{m=1}^{|\gM|}\alpha_{i,m}(\boldsymbol{z}_m\odot\boldsymbol{\theta}^{\delta+\mathtt{pre}}) - \sum_{m=1}^{|\gM|}\alpha_{i,m}(\boldsymbol{z}_m\odot\boldsymbol{\theta}^{\mathtt{pre}})\\
     &= \sum_{m=1}^{|\gM|}\alpha_{i,m}(\boldsymbol{1}\odot\boldsymbol{\theta}^{\mathtt{pre}})  + \sum_{m=1}^{|\gM|}\alpha_{i,m}(\boldsymbol{z}_m\odot\boldsymbol{\theta}^{\delta+\mathtt{pre}}) - \sum_{m=1}^{|\gM|}\alpha_{i,m}(\boldsymbol{z}_m\odot\boldsymbol{\theta}^{\mathtt{pre}})\\
      &= \sum_{m=1}^{|\gM|}\alpha_{i,m}((\boldsymbol{1}-\boldsymbol{z}_m)\odot\boldsymbol{\theta}^{\mathtt{pre}}  + \boldsymbol{z}_m\odot\boldsymbol{\theta}^{\delta+\mathtt{pre}}).
\end{align}
We have assumed that $\sum_{m=1}^{|\gM|}\alpha_{i,m}=1$ which we have ensured through normalizing the expert activation in Alg.~\ref{alg:smat_metatraining}. The result shows that each task model $\boldsymbol{\theta}_i$ can now be interpreted as a weighted sum of different experts, where each expert (i.e.,$(\boldsymbol{1}-\boldsymbol{z}_m)\odot\boldsymbol{\theta}^{\mathtt{pre}}  + \boldsymbol{z}_m\odot\boldsymbol{\theta}^{\delta+\mathtt{pre}})$) is a sparse interpolation between pre-trained $\theta^{\mathtt{pre}}$ and meta-tuned $\theta^{\mathtt{pre}+\delta}$ models in the parameter space.